\pdfoutput=1

\documentclass[11pt]{article}

\usepackage[preprint]{acl}

\usepackage{times}
\usepackage{latexsym}

\usepackage[T1]{fontenc}

\usepackage[utf8]{inputenc}

\usepackage{microtype}

\usepackage{inconsolata}

\usepackage{graphicx}
\usepackage{tikz}
\usepackage{booktabs}
\usepackage{url}

\usetikzlibrary{shapes.geometric, positioning, arrows}

\definecolor{softblue}{RGB}{150, 180, 200}
\definecolor{softgreen}{RGB}{150, 200, 120}
\definecolor{softred}{RGB}{220, 180, 180}

\title{Matching Meaning at Scale: Evaluating Semantic Search for 18th-Century Intellectual History through the Case of Locke}

\author{Yu Wu\textsuperscript{1}, 
  Ananth Mahadevan\textsuperscript{1}\thanks{Contribution made during doctoral studies.}, 
  Filip Ginter\textsuperscript{2,3}, 
  Michael Mathioudakis\textsuperscript{1}, 
  Mikko Tolonen\textsuperscript{1} \\
  \textsuperscript{1}University of Helsinki \quad
  \textsuperscript{2}TurkuNLP, University of Turku \quad 
  \textsuperscript{3}ELLIS Institute Finland\\
  \texttt{\{yu.wu, ananth.mahadevan, michael.mathioudakis, mikko.tolonen\}@helsinki.fi} \\
  \texttt{figint@utu.fi}}

\begin{document}
\maketitle
\begin{abstract}
While digitized corpora have transformed the study of intellectual transmission, current methods rely heavily on lexical text reuse detection, capturing verbatim quotations but fundamentally missing paraphrases and complex implicit engagement. This paper evaluates semantic search in 18th-century intellectual history through the reception of John Locke's foundational work. Using expert annotation grounded in a semantic taxonomy, we examine whether an off-the-shelf semantic search pipeline can surface meaning-level correspondences overlooked by lexical methods. Our results demonstrate that semantic search retrieves substantially more implicit receptions than lexical baselines. However, linguistic diagnostics also reveal a ``lexical gatekeeping'' effect, where retrieval remains partially constrained by surface vocabulary overlap. These findings highlight both the potential and the limitations of semantic retrieval for analyzing the circulation of ideas in large historical corpora. The data is available at \url{https://github.com/COMHIS/locke-sim-data}.
\end{abstract}

\section{Introduction}
The digitization of vast historical archives, such as Eighteenth Century Collections Online (ECCO)~\citep{tolonenAnatomyEighteenthCentury2022} and Early English Books Online (EEBO)~\citep{makelaOpeningBlackBox2025}, together with recent advances in Natural Language Processing (NLP), has created unprecedented opportunities for intellectual history~\citep{comin2026}. Digital methodologies now allow scholars to move beyond the study of isolated authorial intent and begin to “recreate the cultural terrain” of historical debate by tracing how ideas circulated across large textual corpora~\citep{edelsteinIntellectualHistoryDigital2016}. At the same time, this shift from close reading to large-scale computational analysis exposes a central methodological bottleneck: the semantic gap between lexical similarity and meaning-level relationships preserved in textual passages. 

Traditional computational approaches to reception history rely heavily on lexical text reuse detection~\citep{rossonReceptionReaderExploring2023}; for further discussion in the context of intellectual history, see \citet{comin2026}. While effective for identifying verbatim quotations, such approaches struggle with implicit reference, paraphrase, Optical Character Recognition (OCR) errors, diachronic lexical change, and the conceptual complexity of philosophical language. As a result, many forms of intellectual engagement between texts remain difficult to detect at scale.



One context in which this semantic gap becomes particularly visible is 18th-century British intellectual history. The period is widely recognised as a locus of Enlightenment contestation and a rapidly expanding print public, where arguments about reason, religion, science, and politics circulated across genres and decades~\citep{porter2001enlightenment}. Tracing how such arguments were reformulated and redeployed is essential for understanding the transmission of ideas, yet doing so across millions of publications is beyond the scope of manual scholarship. While recent advances in semantic search based on dense vector retrieval~\citep{reimersSentenceBERTSentenceEmbeddings2019, douzeFAISSLibrary2025} make it theoretically plausible to detect semantic similarity beyond exact lexical overlap in the time frame and language domain, their reliability on uncurated historical corpora remains empirically untested. 

To operationalise this inquiry, we focus on John Locke (1632--1704)’s \textit{An Essay Concerning Human Understanding} (1689). Because the \textit{Essay} was published at the end of the seventeenth century, its reception spans the entire 18th-century corpus analysed here, allowing its influence to be traced across the full period. This is not optimal for authors such as David Hume, whose major Enlightenment publications appear only from the late 1730s onwards. Locke’s work is also particularly suitable because his intellectual legacy is more contested and complex than the traditional narrative of the “father of liberalism” suggests~\citep{stantonJohnLockeFable2018}. Recent scholarship has highlighted the fragmented nature of his reception, which developed into distinct republican and libertarian traditions~\citep{laymanLockesRepublicanLiberal2021}. 
At the level of textual practice, Locke’s arguments often appear in structured formulations that are subsequently reformulated or paraphrased in later texts. In such cases, lexical searches fundamentally fail to capture the semantic footprint of the idea.

This paper therefore presents a proof-of-concept evaluation of semantic search for 18th-century historical corpora through the case of Locke’s quotations. We examine whether a minimal, off-the-shelf semantic search setup can identify meaning-level correspondences in the reception of Locke that lexical text reuse detection methods miss. We first operationalize ``reception'' into a hierarchical framework that distinguishes genuine semantic engagement, such as paraphrases and implicit meaning matches, from both verbatim text reuse and mere topical matches. Guided by this framework, we implement a multi-stage evaluation workflow, annotating and analyzing iteratively in non-lexical hits of our semantic search pipeline. Based on the annotated subset, our mixed-methods analysis demonstrates that semantic search uncovers an order of magnitude more implicit reception than traditional string matching across core historical queries. Finally, through a linguistic diagnostic encompassing vocabulary overlap, OCR noise, and syntactic divergence, we expose a profound ``lexical gatekeeping'' effect: latent lexical similarities continue to shape retrieval outcomes, even within dense semantic spaces.

To summarize, our main contributions are:
\vspace{-0.5em}
\begin{itemize}
    \itemsep -0.3em
    \item A proof-of-concept evaluation using Locke as a case study, demonstrating that semantic search retrieves substantially more implicit receptions than lexical text-reuse baselines.
    \item A conceptual framework for modelling reception that distinguishes between different contextual matches at the level of semantic relationships between passages.
    \item A reusable iterative evaluation workflow that applies the annotation framework to identify meaning-level receptions and their historical significance.
    \item A linguistic diagnostic analysis revealing a \textit{lexical gatekeeping} effect despite robustness to OCR noise and syntactic variation.
\end{itemize}

\section{Related Work}
\paragraph{Digitized Infrastructure and Lexical Methods}
The mass digitization of historical archives~\cite{armitageHistoryManifesto, WorldFictionDigital2018, langlais_pierre_carl_2021_4751204}, coupled with expansive structured metadata and relational networks~\cite{hillReconstructingIntellectualNetworks2019, bermesTextDataLinkmining2017}, has established the infrastructure for computational intellectual history. Retrieving specific conceptual evidence from these corpora has traditionally relied on lexical methodologies. Techniques ranging from Keyword in Context (KWIC)~\cite{luhnKeyWordincontextIndex1960}, word-level distributional semantics~\cite{fitzmauriceLinguisticDNAInvestigating2017}, and probabilistic topic modeling~\cite{bleiProbabilisticTopicModels2012, goldstoneQuietTransformationsLiterary2014}, to advanced hashing and sequence-alignment algorithms for OCR-resilient text reuse~\cite{mahadevanTextReuseLarge2025, duringImpressoTextReuse2023, vesantoSystemIdentifyingExploring2017, smithDetectingModelingLocal2014}, mark the current limits of surface-level matching. Bridging this persistent ``semantic gap'' requires moving beyond static methods to neural, passage-level models capable of tracking complex conceptual shifts~\cite{mcgillivray10WhyDoes2024, giulianelli-etal-2020-analysing}.

\paragraph{Semantic Search in Historical Corpora}
To advance beyond lexical navigation, digital humanists increasingly use dense Information Retrieval (IR) models, such as Dense Passage Retrieval~\cite{karpukhinDensePassageRetrieval2020} and Sentence Transformer~\cite{reimersSentenceBERTSentenceEmbeddings2019}, to capture compositional semantics, bypassing explicit string matching. Recent applications demonstrate the capacity of bi-encoders to capture complex intertextualities~\cite{schelbLociSimilesBenchmark2026} and semantic parallels~\cite{kanervaSemanticSearchExtractive2025a, franklinNewsDejaVu2024, alrahabiAutomaticTextComparison2025} that remain invisible to lexical searches. 
However, deploying contemporary neural networks on uncurated historical paragraphs exposes severe algorithmic vulnerabilities. Alongside OCR-induced tokenization artifacts~\cite{zhangOCRHindersRAG2025, michailCheapCharacterNoise2025}, temporal semantic degradation~\cite{lazaridouMindGapAssessing2021} and entity-centric popularity biases~\cite{sciavolinoSimpleEntityCentricQuestions2021}, dense models regress to shallow vocabulary matching when confronted with out-of-domain historical syntax~\cite{rizwanResolvingLexicalBias2025, macavaneyABNIRMLAnalyzingBehavior2022}. This limitation is further entrenched by standard IR benchmarks, which structurally penalize true semantic retrieval through lexically pre-filtered ground truth~\cite{thakurBEIRHeterogeneousBenchmark2021}. 
Therefore, rigorously evaluating semantic architectures within real-world historical corpora is critical to distinguish genuine meaning retrieval from lingering algorithmic biases. 

\paragraph{Computational Early Modern Reception}
Early computational methods successfully mapped early modern intellectual networks using structured metadata from social and epistolary graphs~\cite{warrenSixDegreesFrancis2016a, edelsteinHistoricalResearchDigital2017, burrowsFrenchBookTrade2012}, while lexical pipelines tracked verbatim text reuse~\cite{roeTextReuseCultural2024, tolonenReceptionDavidHumes2025} and extracted explicit quotations~\cite{muznyTwostageSieveApproach2017, brunnerCorpusREDEWIEDERGABE2020}. 
However, the highly contested intellectual legacy of foundational figures like John Locke~\cite{collinsShadowLeviathanJohn2020, hanckLockesConfusionConfused2019} exemplifies the requirement of moving to passage-level semantic mappings: his foundational concepts were widely appropriated without explicit citation, relying on implicit references and shifting vocabularies that completely evade traditional lexical searches~\cite{harrisOriginGovernmentAfterlives2023, careyLockeShaftesburyHutcheson2006}. Using Locke’s legacy as an empirical stress test, our study evaluates whether contemporary semantic retrieval can bypass surface-level lexical biases to uncover hidden conceptual reuse within the noisy 18th-century archive.

\section{Evaluation Methodology}
\label{sec:method}
\begin{figure*}[t]
    \centering
    \small
    

    \tikzstyle{block} = [rectangle, draw=black!60, fill=gray!10, 
        text width=6.5em, text centered, rounded corners, minimum height=2.5em, line width=0.8pt]
    \tikzstyle{semblock} = [rectangle, draw=black!60, fill=softblue, 
        text width=6.5em, text centered, rounded corners, minimum height=2.5em, line width=0.8pt]
    \tikzstyle{humanblock} = [rectangle, draw=black!60, fill=softred, 
        text width=6.5em, text centered, rounded corners, minimum height=2.5em, line width=0.8pt]
    \tikzstyle{decision} = [diamond, draw=black!60, fill=yellow!10, aspect=2, 
        text centered, inner sep=1pt, line width=0.8pt, font=\footnotesize]
    \tikzstyle{line} = [line width=0.8pt, -triangle 45, draw=black!70]

    \begin{tikzpicture}[node distance=1.2cm and 1.5cm, auto]
        
        \node [block] (quote) {Target Locke Quotes};
        
        \node [block, above right=0cm and 1.2cm of quote] (blast) {BLAST Lexical Hits};
        \node [semblock, below right=0cm and 1.2cm of quote] (sem) {FAISS Semantic Hits};
        
        \node [block, right=5cm of quote] (filter) {Anti-Lexical Filter};
        
        \node [block, above right=0.2cm and 2cm of filter] (uniqblast) {Unique Lexical Hits};
        \node [block, right=2cm of filter] (intersection) {Intersection};
        \node [semblock, below right=0.2cm and 2cm of filter] (uniqsem) {Unique Semantic Hits};
        
        \draw[line] (quote.east) -- (blast.west);
        \draw[line] (quote.east) -- (sem.west);
        
        \draw[line] (blast.east) -- (filter.west);
        \draw[line] (sem.east) -- (filter.west);

        \draw[line] (filter.east) -- (uniqblast.west);
        \draw[line] (filter.east) -- (intersection.west);
        \draw[line] (filter.east) -- (uniqsem.west);

        \node [humanblock, below=0.6cm of uniqsem] (sample) {Stratified Sampling};
        
        \draw[line] (uniqsem.south) -- node[right, font=\scriptsize] {Initial} (sample.north);

        \node [humanblock, below=0.6cm of sample] (annotate) {Expert Annotation};
        \node [humanblock, left=of annotate] (eval) {Evaluate Density};
        \node [decision, left=of eval] (judge) {Sufficient Signal?};
        \node [humanblock, left=of judge] (end) {Historical Analysis};

        \draw[line] (sample.south) -- (annotate.north);
        \draw[line] (annotate.west) -- (eval.east);
        \draw[line] (eval.west) -- (judge.east);
        
        \draw[line] (judge.west) -- node[above, font=\scriptsize] {Yes} (end.east);
        
        \draw[line] (judge.north) |- node[pos=0.2, right, font=\scriptsize] {No} node[pos=0.8, above, font=\scriptsize] {Iterative Resampling} (sample.west);
        
    \end{tikzpicture}
    
    \caption{The overall evaluation workflow. The automated search pipeline (top) extracts and filters semantic candidates (blue blocks) against the lexical baseline. The human-in-the-loop workflow (bottom, red blocks) systematically annotates the resulting \textit{Unique Semantic Hits} to evaluate signal density and historical significance.}
    \label{fig:architecture}
    
    \vspace{-1em}
\end{figure*}
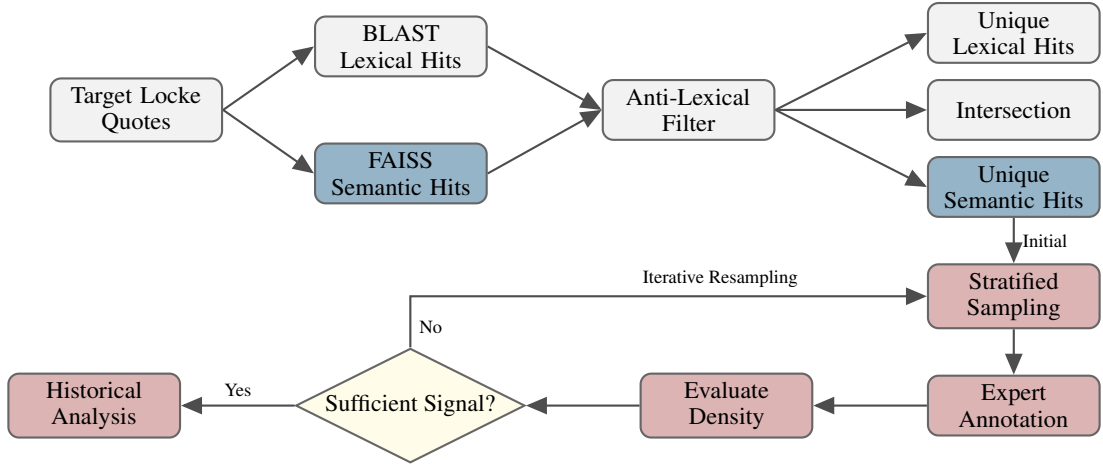 

Our evaluation framework is designed to isolate significant meaning matches from mere lexical reuse. By conceptualizing the retrieval hits as potential reception signals, we aim to uncover their underlying matching distribution and the analytical conditions under which a semantic match constitutes robust historical evidence.

\subsection{Heuristic Annotation Taxonomy}
\label{sec:labels}

To ground the iterative workflow in a stable interpretive framework, we introduce an annotation taxonomy for modelling reception that distinguishes between different types of textual correspondence. The framework is designed as an operational heuristic to facilitate rapid triage rather than imposing a rigid typology. It is a hierarchy of heuristic matches consisting of \textit{Lexical Matches}, \textit{Paraphrase Matches} (\textit{Paraphrases} for short), and \textit{Meaning Matches}, alongside a separate category of \textit{Topical Matches}. \textit{Lexical Matches} capture verbatim reuse, while \textit{Paraphrase Matches} identify cases in which wording has changed but a clear quotation relationship remains. \textit{Meaning Matches}, by contrast, refer to passages that display semantic engagement suggesting that one text meaningfully interacts with another even without explicit lexical overlap. \textit{Topical Matches}, in turn, refer to passages that address the same subject without semantic evidence of such engagement. For cases of OCR failure or total semantic divergence, we supplement this hierarchy with a \textit{No Match} category, and \textit{Don't Know} for the cases where the semantic connection is too ambiguous for rapid judgment, requiring extensive close reading to verify. In our evaluation, \textit{Paraphrases} and \textit{Meaning Matches} are aggregated as \textit{significant hits}, representing genuine intellectual reception.

\subsection{Iterative Evaluation Strategy}
\label{sec:iter_eval}
To extract high-signal reception from a noisy retrieval pool, we formalize a multi-stage, human-in-the-loop evaluation framework, illustrated as Fig.~\ref{fig:architecture}. We begin by sampling highly reused quotes from Locke's texts to serve as our search queries. We then deploy a pilot sampling strategy across the broad retrieval pool to understand the basic semantic distribution without presupposing algorithmic success. Finally, we conditionally target our deepening annotation based on these initial results. This progressive annotation process optimizes expert bandwidth, prioritizing genuine reception clusters over algorithmic noise.

\subsection{Analytical Framework}
\label{sec:analytic}
To evaluate the pipeline's utility and its underlying constraints, our framework covers content-level statistical analysis and linguistic-level diagnosis. Firstly, we evaluate the retrieval performance by analyzing how the annotated semantic labels and metadata are distributed across the ranking. Secondly, to interpret the linguistic patterns identified in these results, we first map the annotated English hits into a standard confusion matrix (TP, TN, FP, and FN subsets) based on their ranking intervals and semantic validity. Next, within each quadrant of the matrix, we compute three targeted linguistic features to decode what triggers these specific errors or successes: \textbf{Vocab Sim.} (Jaccard similarity of word lemmas) quantifies the model's dependency on surface-level lexical anchoring, while \textbf{OOV} (Out-of-Vocabulary rate on query and target separately) serves as a proxy for historical OCR noise. Additionally, \textbf{POS Div.} utilizes Jensen-Shannon divergence over Part-of-Speech distributions to measure linguistic structural rephrasing independent of vocabulary.


\section{Empirical Implementation}
This section details the experimental setup and evaluation workflow operated based on our evaluation methodology described in Sec.~\ref{sec:method}. 

\subsection{Corpus}
Our corpus is the official OCR-extracted text provided by the Eighteenth Century Collections Online (ECCO)~\citep{tolonenAnatomyEighteenthCentury2022}, which contains over 32 million book pages from the 18th century. This dataset serves as an ideal testbed for tracing John Locke’s reception, as his influence permeates the period’s diverse genres, from political tracts to educational manuals. 
The comprehensive corpus undergoes chunking (as detailed in Sec.~\ref{sec:pipeline}) to build the core index for the vectorized search. 
To manage duplicate editions and ensure bibliographic sufficiency, we integrated metadata from the English Short Title Catalogue (ESTC), specifically the work IDs, authors, publication years of different versions, and genres. The genre labels are assigned based on the Genre Classification Workflow developed by~\citet{tiihonenGenreClassificationWorkflow2025}. 

\subsection{Text Reuse Baseline}
As our primary baseline for capturing explicit intellectual reception, we utilize the comprehensive text reuse network generated across the ECCO corpus~\citep{rossonReceptionReaderExploring2023}. This infrastructure relies on an adaptation of BLAST (Basic Local Alignment Search Tool). Its fuzzy-matching capabilities inherently tolerate 18th-century orthographic variations and persistent OCR noise, effectively mapping any text input to verbatim quotations, exact reprints, and lightly edited excerpts across millions of pages. Despite the limitation on semantic-level matching, we repurpose the lexical reuse for three foundational tasks: establishing a comparative baseline, sourcing evaluation queries (Sec.~\ref{sec:query}), and purifying our unique semantic hits (Sec.~\ref{sec:pipeline}).

\subsection{Query Source and Construction}
\label{sec:query}
In alignment with grounding computational evaluation in 18th-century British intellectual history, we selected John Locke’s \textit{An Abridgment of Mr Locke’s Essay Concerning Human Understanding} (ESTC R22993) as a highly representative, scalable case study. Published prior to the 18th century, this foundational work allows us to track intellectual reception across the entirety of the ECCO timeline. Because its historical significance is universally recognized and it possesses a dense, well-documented network of existing lexical reuses, it provides an ideal, undisputed baseline without the need to overstate the novelty of the source itself. 

To capture genuine external lexical reuse rather than isolated fragments, we queried the baseline for reuse text clusters originating from this work, as introduced by \citet{rossonReceptionReaderExploring2023}. In this pipeline, we formally define a \textit{quote} not as a punctuated citation, but as a contiguous, verifiable fragment of reused text. We isolated these quotes by enforcing a length constraint of 150–300 characters, and a minimum external reuse frequency (n $\geq$ 3)\footnote{These thresholds were empirically determined to eliminate uninterpretable short snippets and overly long, multi-page sequences that obscure targeted semantic analysis.}. 

\subsection{Minimal Semantic Search Pipeline}
\label{sec:pipeline}

We adopt a deliberately minimal dense retrieval-based search pipeline to evaluate the conceptual semantic capabilities of semantic search. The implementation of core infrastructure is publicly accessible via the \texttt{semantic-sim} repository\footnote{\url{https://github.com/TurkuNLP/semantic-sim}}. The underlying vector index was built upon a broader Early Modern English corpus (encompassing both 18th-century ECCO and pre-18th-century EEBO texts). We can easily isolate the target ECCO corpus by filtering the retrieved outputs via document IDs. 
The texts were segmented into 100-token chunks and encoded using the \texttt{paraphrase-multilingual-mpnet-base-v2} model~\citep{reimersMakingMonolingualSentence2020, songMPNetMaskedPermuted2020}. This encoder natively supports over 50 languages, including English, French, Italian, and Spanish, while allowing us to evaluate its zero-shot capability on unsupported historical languages such as Latin. Efficient vector indexing and retrieval were executed via FAISS (Facebook AI Similarity Search)~\citep{douzeFAISSLibrary2025}. For each query, this index outputs a ranked list of retrieved segments with their semantic similarity scores to the input. We referenced these ranked outputs against our BLAST lexical baseline. By computing the set difference between the two systems, we filtered out known verbatim overlap.

\subsection{Annotation Setup}
\begin{figure}[t]
\begin{center}
\includegraphics[width=0.9\columnwidth]{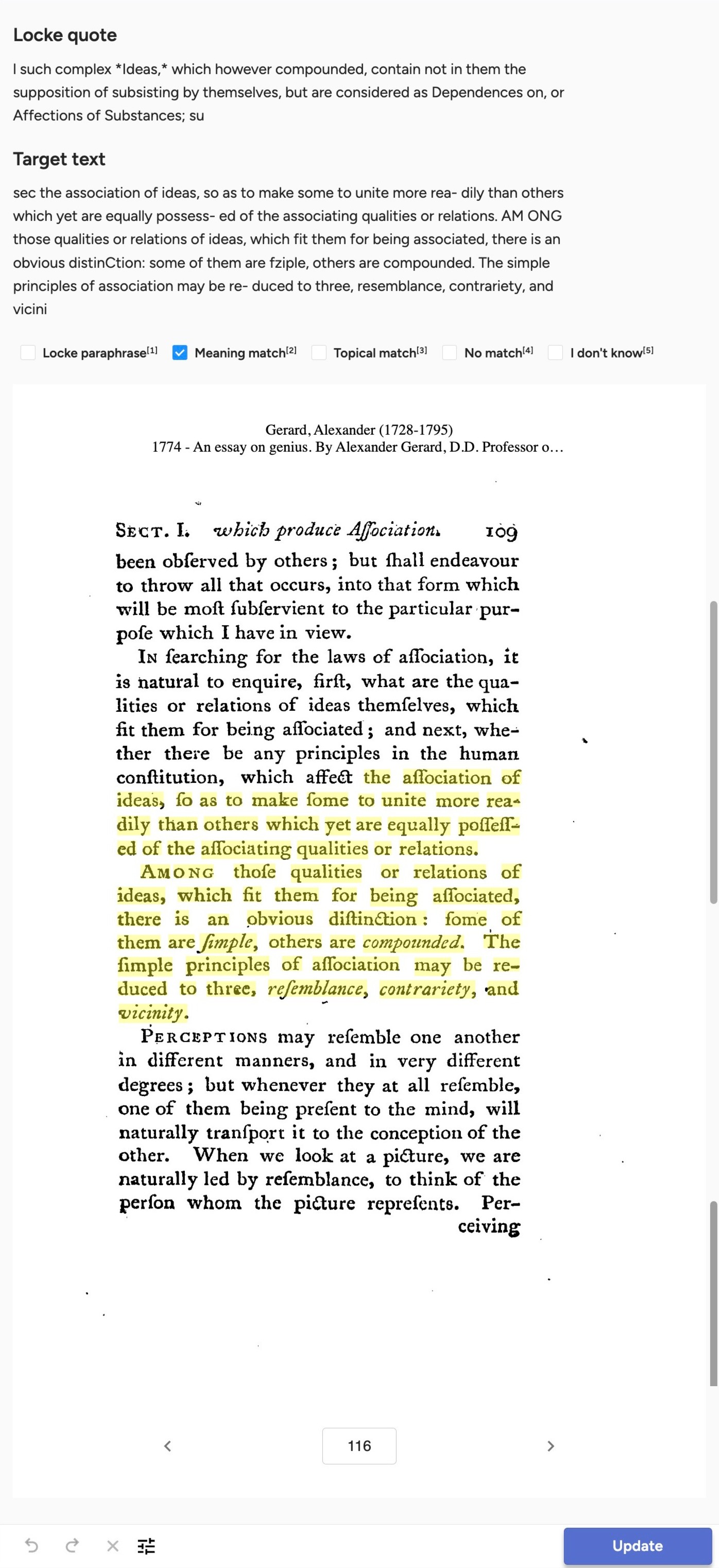}
\caption{The annotation interface screenshot.}
\label{fig:interface}
\end{center}
\vspace{-1.5em}
\end{figure}

We deployed a project in the locally hosted instance of Label Studio~\citep{LabelStudio} for the manual annotation. As illustrated in Fig.~\ref{fig:interface}, the interface was designed to prevent annotators from evaluating fragments in a vacuum. Alongside the retrieved chunk and Locke's original quote, the UI simultaneously displayed the corresponding primary author, work title, and publication year, and provided direct access to the broader document context. The annotation was conducted by five domain experts from the Computational History Group (COMHIS) at University of Helsinki. Each candidate was assigned to a single annotator for rapid triage. Ambiguous cases were marked as \textit{Don’t Know} and processed independently.

\subsection{Annotation Process}
We implemented a workflow following our methodology in Sec.~\ref{sec:method} and all the setup above to deploy our iterative evaluation strategy. The empirical outcomes and the evidence for iterative deepening are detailed in Sec.~\ref{sec:result}.

\paragraph{Source Selection}
From the top 1,000 most frequently reused segments extracted in Sec.~\ref{sec:query}, we selected the top-5 highest-frequency quotes, supplemented by 5 randomly sampled from each of three subsequent frequency tiers (ranks 6-50, 51-150, and 151-1,000), yielding 20 search queries.

\paragraph{Search and Filtering}
For these 20 queries, we retrieved the top 200,000 semantic hits using our pipeline established in Sec.~\ref{sec:pipeline}. Filtering by document ID retained an average of 173,266 ECCO hits per query (86.6\%). To prevent the search space from being dominated by intra-document repetition or frequent reprints, we deduplicated the results by ESTC work ID. Retaining only the highest-scoring hit per distinct work yielded an average of 13,391 unique candidates per query. We then filtered out known verbatim overlap.

\paragraph{Stratified Pilot Sampling}
We conducted a probing manual annotation of 50 hits per quote, sampling the top 5 results and uniformly distributed intervals across the top 90\% of the filtered hits. They were distributed among five domain experts and annotated according to the taxonomy in Sec.~\ref{sec:labels}. 

\paragraph{Iterative Deepening}
We annotated more top hits for specific quotes according to a dual criterion: a high density of semantic matches and substantive historical significance. We set the top 200 as an empirical annotation upper bound. Before the exhaustive annotation, we annotated 50 samples within the top-200 hits for a quick triage to verify that the promising yields from our initial broad but sparse sampling were not merely algorithmic outliers. They were the top 20 hits, plus 30 hits sampled iteratively (one every six) from the remainder. By confirming a consistently dense semantic signal first, we mitigated the risk of exhaustively annotating a noise-heavy set that would force us to restart the deepening process. Ultimately, candidates maintaining robust yields triggered the exhaustive annotation of their complete top-200 results for deep-dive historical analysis.

\section{Results and Analysis}
\label{sec:result}
Our analysis assesses the reliability and technical limitations of the semantic search pipeline.

\subsection{Overall Results}
\begin{table*}[t]
\centering
\small
\begin{tabular}{l|c|ccc|c}
\toprule
\textbf{Category} & \textbf{Overall (\%)} & \textbf{In Top-5 Ranked(\%)} & \textbf{In Top 20\% (\%)} & \textbf{In Top 50\% (\%)} & \textbf{Spearman's $\rho$} \\
\midrule
Paraphrase & 2.1 & 18.0 & 7.0 & 3.5 & $-0.224$*** \\
Meaning Match & 11.6 & 44.0 & 26.3 & 16.3 & $-0.294$*** \\
Topical Match& 31.1 & 24.0 & 34.0 & 32.7 & $-0.054$\phantom{***} \\
No Match & 54.8 & 14.0 & 32.3 & 47.0 & 0.307*** \\
Don't Know & 0.4 & 0 & 0.3 & 0.5 & $-0.026$\phantom{***} \\
\bottomrule
\end{tabular}
\caption{Distribution and ranking validity of semantic categories. Overall, Top-5 Ranked, Top 20\%, and Top 50\% yields are reported as mean percentages across the sampled hits, calculated based on original retrieval rankings. For each category, Spearman's rank correlation ($\rho$) is calculated between its binary indicator and the local retrieval rank within the sampled subset of each query. Significance levels: *** $p < 0.001$.}
\label{tab:overall_stats}
\vspace{-1em}
\end{table*}

\begin{figure}[t]
\begin{center}
\includegraphics[width=\columnwidth]{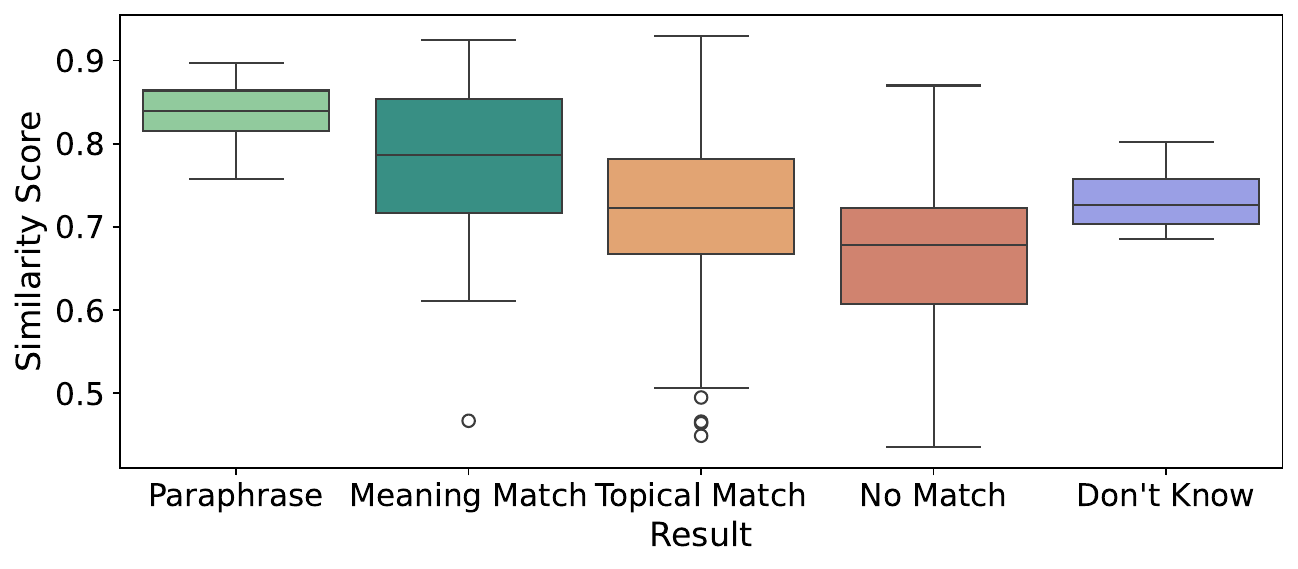}
\caption{Similarity scores across categories, indicating a lack of global calibration. }
\label{fig:overall_box}
\end{center}
\vspace{-1.5em}
\end{figure}

Before detailing the annotation outcomes, we quantify the pipeline's lexical filtering efficacy. Semantic search achieved a 95.5\% recall of BLAST exact matches (averaging 103.9 intersection vs. 4.9 unique BLAST hits), confirming it effectively matched OCR-resilient lexical baselines even before capturing semantic variance. 

On this highly refined dataset, our heuristic taxonomy facilitated an efficient rapid-triage evaluation, logging a median annotation time of approximately 19.3 seconds per candidate. The resulting annotations demonstrate that the model possesses a robust ordinal capability. As detailed in Tab.~\ref{tab:overall_stats}, the ranking validity, measured by Spearman's rank correlation ($\rho$), exhibits a highly significant monotonic relationship with human judgments. The most informative semantic categories (\textit{Paraphrase} and \textit{Meaning Match}) show strong negative correlations with the hit ranks, confirming that the system surfaces substantive semantically matched fragments to the top of the ranking. This efficacy is most pronounced at the very top: the top-5 ranked hits are predominantly valid semantic matches, with zero of \textit{No Match} across 15 of the 20 quotes.


\begin{figure}[t]
\begin{center}
\includegraphics[width=\columnwidth]{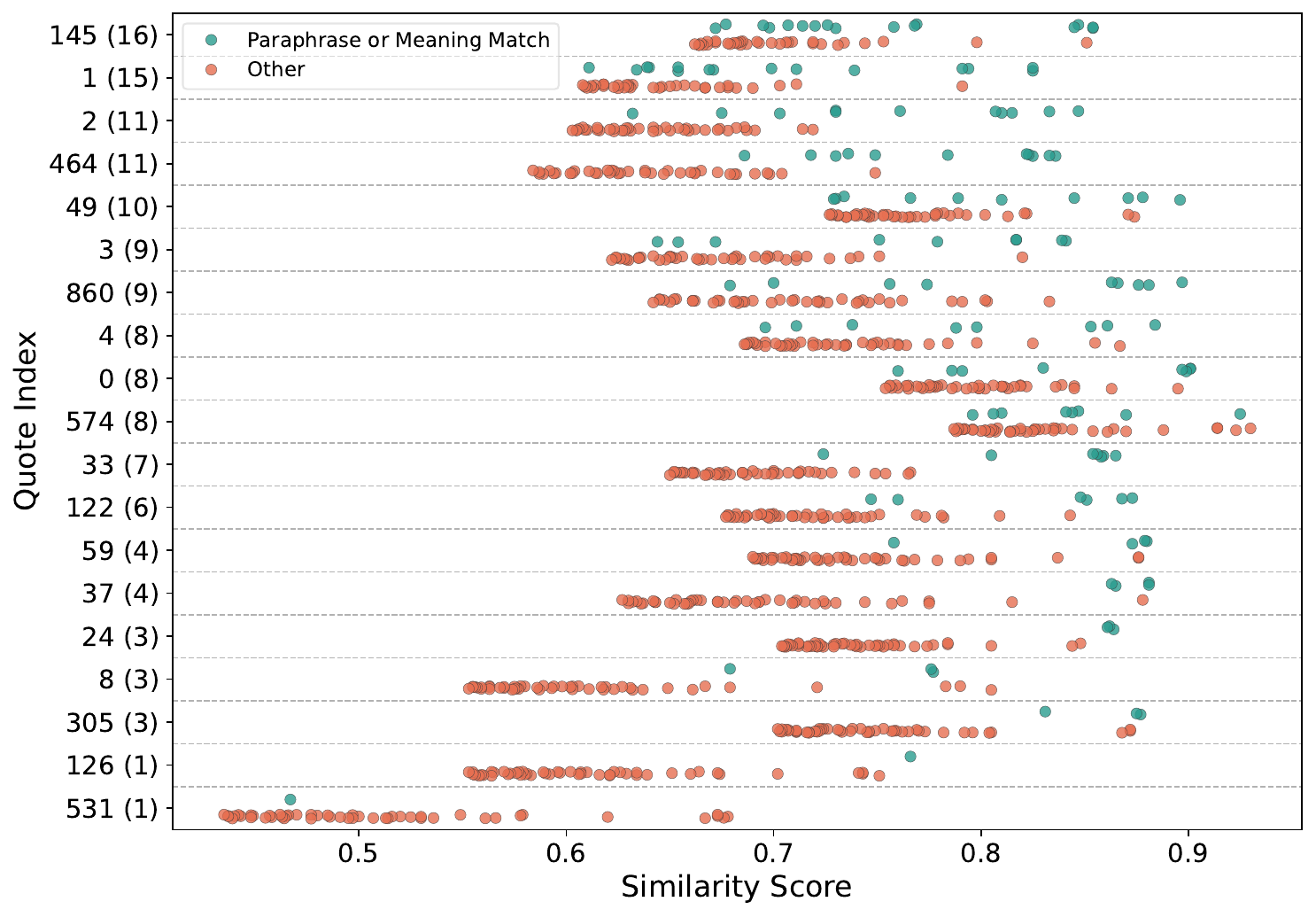}
\caption{Hit distribution per quote, highlighting query-specific heterogeneity. Queries are sorted by the absolute count of significant hits, indicated in parentheses.}
\label{fig:scatter}
\end{center}
\vspace{-1.8em}
\end{figure}

While simply extracting the top-$K$ results suffices for standard ad-hoc retrieval, corpus-level intellectual history requires the complete boundaries of a complete semantic match cluster to quantify macro-level reception patterns. Our evaluation reveals that automating this boundary-drawing is fundamentally bottlenecked by score calibration. As shown in Fig.~\ref{fig:overall_box}, while relative score ranking is correlated to semantic categories, absolute similarity scores overlap significantly across them, preventing the use of a universal global threshold. 

This uncalibrated nature extends deeply to the query level. Fig.~\ref{fig:scatter} demonstrates pronounced query-specific heterogeneity: the score thresholds for significant hits shift drastically depending on the specific query content. More critically, even for individual quotes, these local boundaries remain highly fuzzy, preventing the extraction of a cleanly separated match pool for each quote. While this noise partially reflects the fundamental ambiguity of reception research, where the line between a \textit{Meaning Match} and a mere \textit{Topical Match} is often epistemologically contested, it makes individual thresholds hard to set. This failure of both global and local calibration necessitates a methodological pivot. To determine whether semantic search can still provide inspiring insights for intellectual history, we shift from thresholding strategies to targeted case studies. We prioritize specific quotes with clearer empirical score boundaries to assess whether their high-ranking hits yield substantive meaning matches with historical significance.

\subsection{Individual Results}
\begin{figure}[t!]
    \centering
    \includegraphics[width=\linewidth]{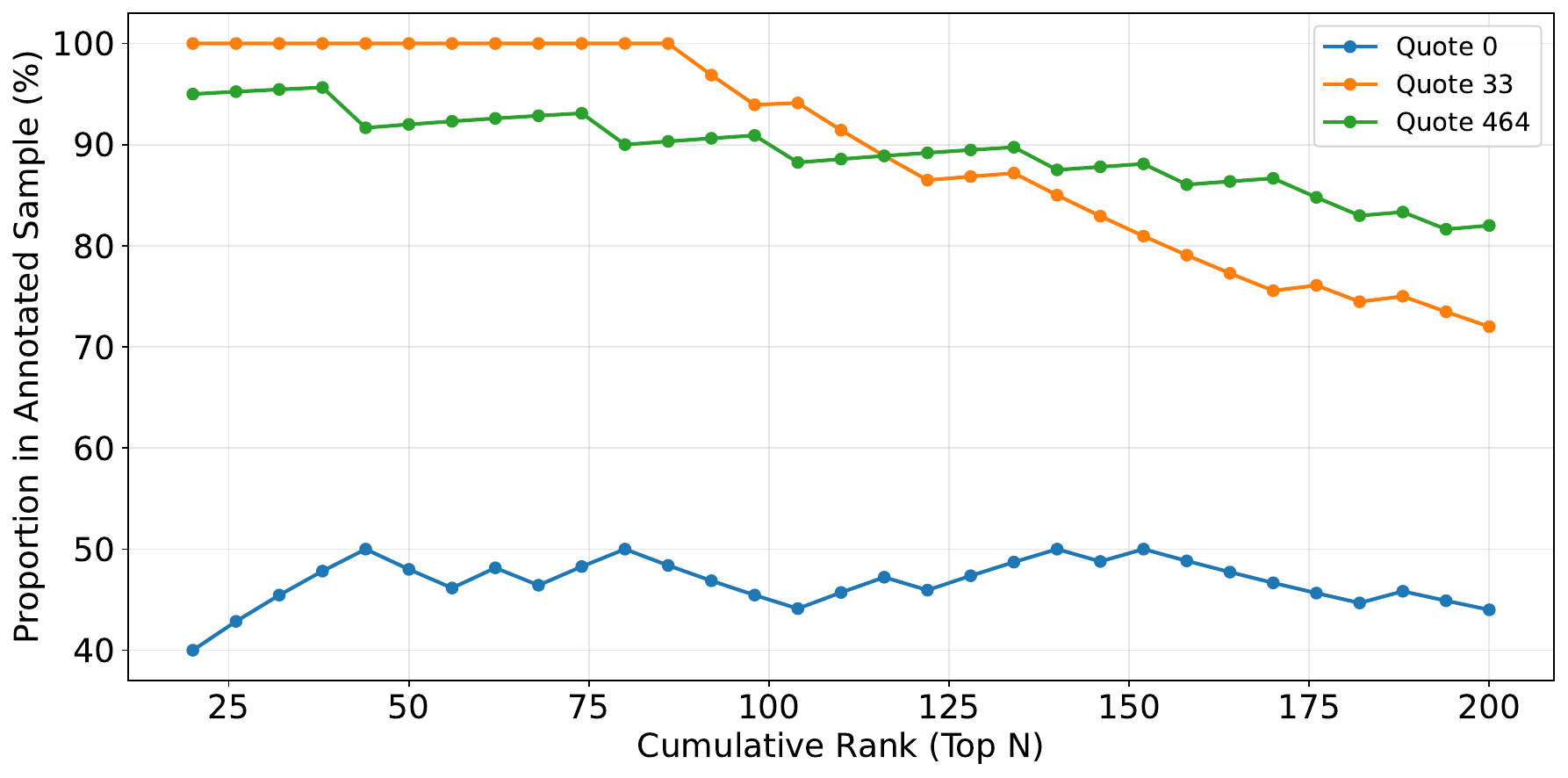}
    \caption{Proportion of significant hits within the 50 validation samples across the top 200 hits for each quote.}
    \label{fig:curve}
    \vspace{-1em}
\end{figure}



To operationalize our iterative deepening (Section~\ref{sec:method}), we finalized three core queries from an initial mixed pool by balancing algorithmic clarity with historical significance. Quotes 33 and 464 were selected as primary anchors because of their clear semantic boundaries at high similarity scores ($>0.8$) alongside profound historical relevance. Conversely, Quote 0, supplemented by domain experts due to its massive lexical footprint, was retained as a comparative baseline. Contrasting with the clear signals of the primary anchors, Quote 0 serves as a rigorous sanity check to verify whether its top-ranked hits reproduce the boundary fuzziness observed during the broad initial sampling. 




 Before committing to the labor-intensive exhaustive annotation of their top hits, we utilized our 50-sample validation set in top-200 hits as a rapid validation mechanism. Fig.~\ref{fig:curve} tracks the proportion of significant hits across this sampling continuum. While the proportion for Quote 0 keeps low in top hits, confirming its boundary fuzziness, Quotes 33 and 464 maintain robust yields above 70\% and 80\%, respectively. This sustained density verifies that these top hits represent deeply cohesive semantic clusters rather than anomalous spikes.

\begin{figure}[t]
\begin{center}
\includegraphics[width=\columnwidth]{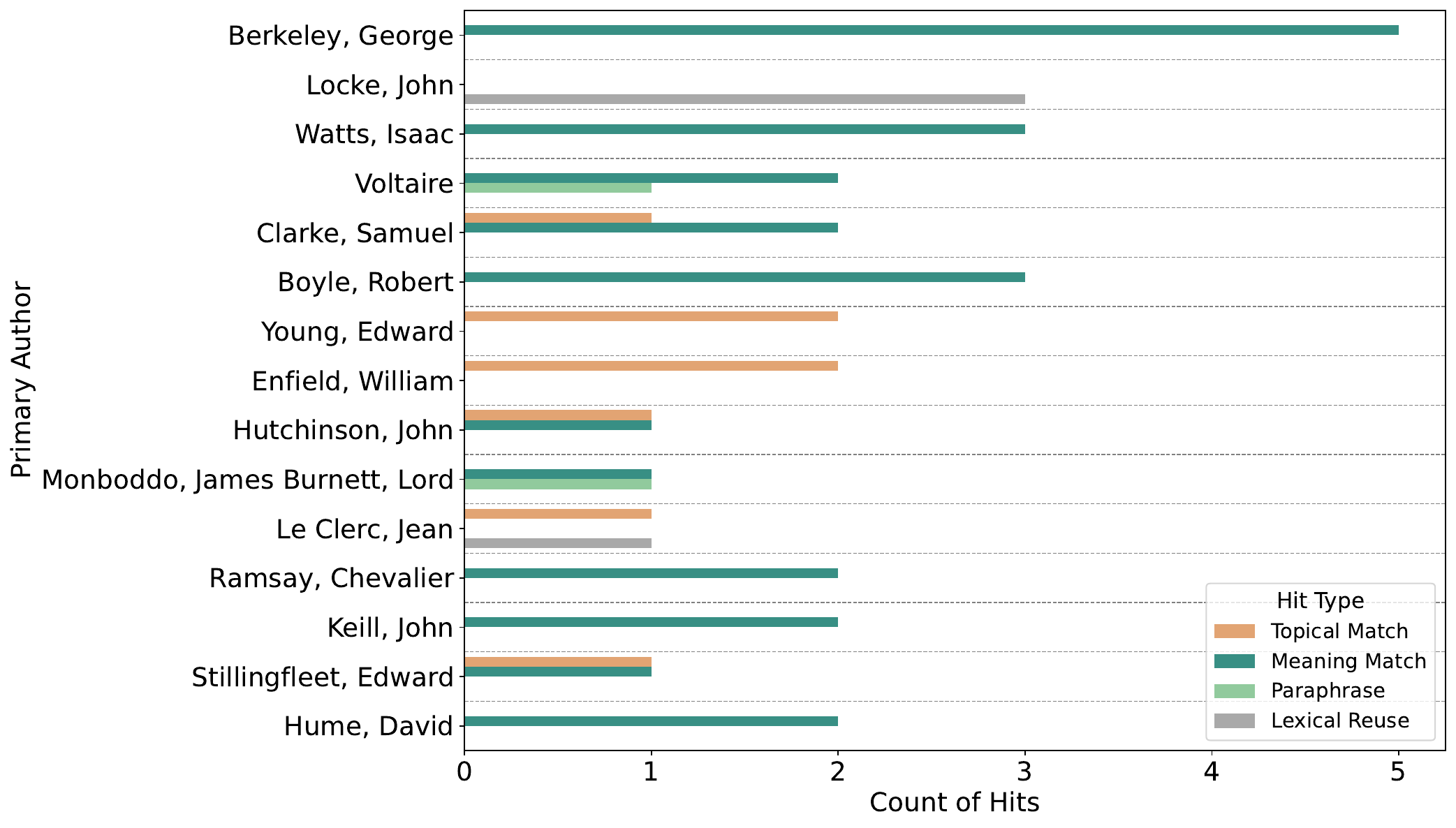}
\caption{Top 15 authors of the top hits for Quote 464.}
\label{fig:quote464_author}
\end{center}
\vspace{-1.5em}
\end{figure}

\begin{figure*}[t]
    \centering
    \includegraphics[width=\textwidth]{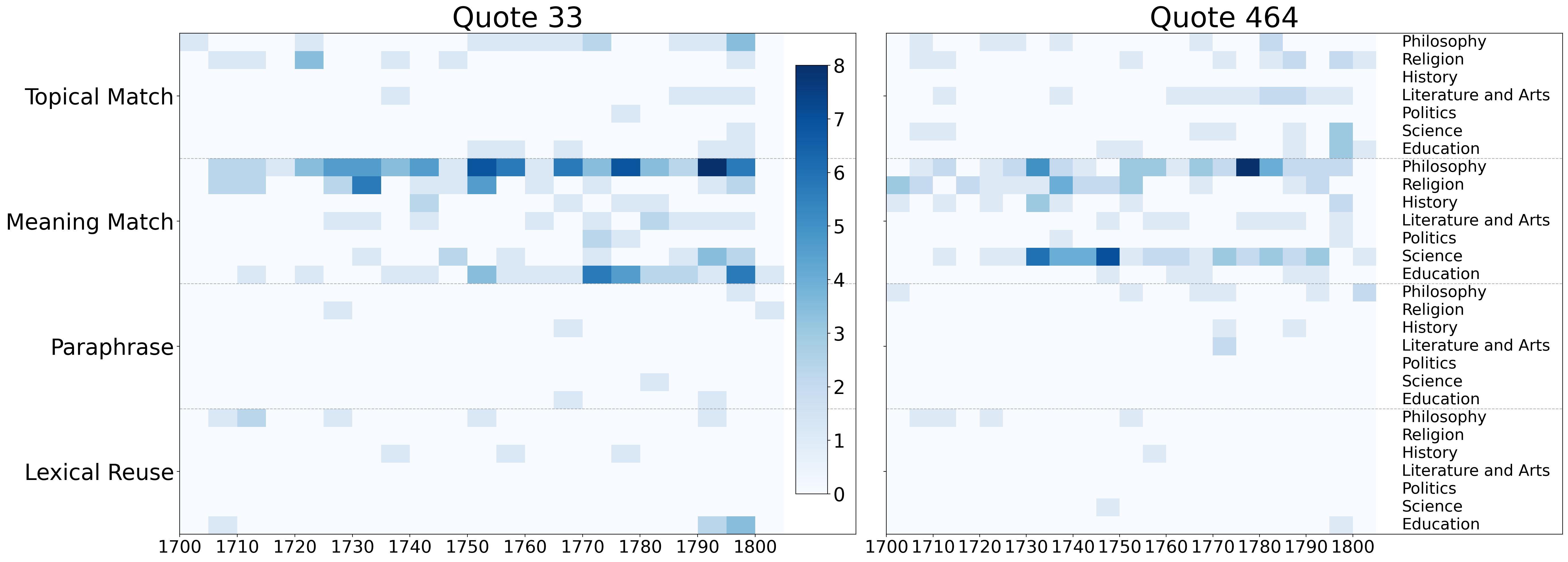}
    \caption{Temporal and disciplinary count distribution of retrieved hits for Quotes 33 (left) and 464 (right). These heatmaps visualize the hits in semantic categories across the seven most dominant 18th-century book genres.}
    \label{fig:quotes_evol}
    \vspace{-1em}
\end{figure*} 

Following the curation of the expert-validated dataset with 400 hits from the two core quotes, we conducted a structural analysis of their historical distribution. A fundamental finding is the sheer volume disparity: significant semantic reuse (\textit{Paraphrase} and \textit{Meaning Match}) outnumbers verbatim lexical reuse by over an order of magnitude (313 vs. 22 hits at the work level), effectively broadening the observable horizon of reception studies. This expansion is starkly visible at the authorial level. Taking Quote 464 as a representative example (Fig.~\ref{fig:quote464_author}), traditional lexical matching predominantly isolates citations by Locke; in contrast, semantic search uncovers a dense, previously invisible network of engagement from central intellectual figures, including Berkeley, Boyle, Hume, and Voltaire.

Projecting these results onto temporal and disciplinary axes (Fig.~\ref{fig:quotes_evol}) reveals that the distribution of reuse types varies heavily by genre. For instance, semantic reuse is significantly more prevalent in \textit{Philosophy} and \textit{Religion} than verbatim quotation. Temporal analysis further reveals distinct trajectories, with different hit types peaking at different decades, suggesting evolving modes of engagement over time. Furthermore, the two analyzed quotes exhibit highly divergent reception profiles: one forms massive, tightly-knit clusters within philosophical and educational discourse, while the other displays a pronounced volume of reuses across the sciences. This macro-level analysis demonstrates the search system's utility in capturing structural intellectual shifts invisible to traditional lexical detection. How these divergent profiles manifest in actual rhetorical repurposing will be the focus of our qualitative case studies.

\subsection{Case Study}

\begin{table}[t]
\centering
\small
\begin{tabular}{@{}lp{0.85\linewidth}@{}}
\toprule
\textbf{ID} & \textbf{Original Quote Fragment for Query} \\ \midrule
33 & l such complex *Ideas,* which however compounded, contain not in them the supposition of subsisting by themselves, but are considered as Dependences on, or Affections of Substances; su \\ \addlinespace[0.5em]
464 & ing their Causes, or any otherwise. The *Atomists,* who define Motion to be *a passage from one place to another,* What do they more than put one synonymous Word for another? For what is *Passage* other than *Motion \\ \bottomrule
\end{tabular}
\caption{Representative quotes for the qualitative study.}
\label{tab:quotes}
\vspace{-1.5em}
\end{table}

To interpret these macro-patterns, we examined the intersection of genre and publication year through targeted close reading. Tab.~\ref{tab:quotes} presents the original texts for these cases. Focusing on Quote 464, from the right sub-figure of Fig.~\ref{fig:quotes_evol}, we observe a distinct peak in reuses within the science category during the 1740s. Qualitative analysis reveals that these hits largely originate from scientific textbooks and manuals. This phenomenon appears closely linked to the consolidation of Newtonianism and the concurrent rebuttal of Cartesian physics, where Lockean epistemological framework concerning motion was repurposed to ground scientific empiricism.

In contrast to this polemical weaponization of Quote 464, Quote 33 illustrates an entirely different trajectory of conceptual drift: pedagogical formalization. As indicated by its dense clustering within the philosophy and education genres, Locke’s complex epistemological framework was systematically adapted into instructional manuals. Rather than engaging with the philosophical nuances, these texts typically simplified and repackaged his core logic to serve didactic purposes.

Crucially, across both divergent trajectories, the 18th-century authors rarely quoted the original sources verbatim. Whether paraphrasing physics for scientific pedagogy or conceptually flattening Locke for educational manuals, this ``implicit reception'' is successfully captured by semantic search. By retrieving these highly reformulated iterations, the semantic search demonstrates its applicability in tracking ideas as they cross disciplinary boundaries and registers of complexity, uncovering context-dependent intellectual shifts that traditional lexical searches would inevitably overlook.

\subsection{Linguistic Diagnostics}

\begin{table}[t]
\centering
\small
\setlength{\tabcolsep}{2pt} 
\resizebox{\columnwidth}{!}{
\begin{tabular}{l|rrrrr}
\toprule
\textbf{Category (\%)} & \textbf{English} & \textbf{French} & \textbf{Latin} & \textbf{Italian} & \textbf{Spanish} \\
\midrule
ECCO Tokens & 92.2 & 3.3 & 3.4 & 0.4 & 0.1 \\
\midrule
Paraphrase & 100.0 & 0.0 & 0.0 & 0.0 & 0.0 \\
Meaning Match & 90.5 & 9.5 & 0.0 & 0.0 & 0.0 \\
Topical Match & 91.0 & 7.4 & 0.0 & 1.3 & 0.3 \\
No Match & 92.7 & 6.2 & 0.9 & 0.2 & 0.0 \\
Don't Know & 100.0 & 0.0 & 0.0 & 0.0 & 0.0 \\
\bottomrule
\end{tabular}
}
\caption{Language distribution across semantic categories in the 1000 pilot samples. The ECCO token language distribution baseline is calculated via token-level aggregation from the ECCO metadata.}
\label{tab:language}
\end{table}

While individual analyses and case studies validate the pipeline's potential, our linguistics-driven analysis reveals the constraints and properties of this success, based on the method proposed in Sec~\ref{sec:analytic}. 

Starting at the macro scale, language identification via Lingua~\cite{pemistahl2021lingua} reveals that the semantic search demonstrates a pronounced capacity for cross-lingual alignment, as detailed in Tab.~\ref{tab:language}. When compared against the token-weighted corpus baseline, the highly relevant ``Meaning Match'' category exhibits a nearly three-fold enrichment of French texts (9.5\%). This non-random distribution across semantic tiers confirms that the multilingual embedding space effectively maps English philosophical propositions onto Continental discourse, even under noisy OCR'ed text, bypassing the need for explicit literal translation. 
Conversely, the model fails to achieve zero-shot semantic transfer for unsupported historical languages like Latin (3.4\% of the corpus), which appears exclusively as noise within the \textit{No Match} category (0.9\%).

However, this cross-lingual reach introduces a specific structural side effect. A close inspection of the retrieval outcomes reveals that the top-ranked FP contains a noticeably higher proportion of non-English hits (14.0\%) compared to the TP (10.0\%). This inflation indicates that the model occasionally exhibits a cross-lingual over-eagerness, elevating non-English texts based on coarse-grained topical proximity rather than strict conceptual equivalence, thereby introducing systemic noise at the head of the search results.

To isolate the underlying model mechanics from this cross-lingual noise, evaluating the purely English subsets exposes rigid monolingual bias, processed using spaCy's large English model~\cite{honnibal2020spacy}. As illustrated by the linguistic diagnostics in Tab.~\ref{tab:linguistic}, TPs inherently rely on a higher degree of lexical overlap with the original query, even though the proportion is very small, with a maximum of 0.030. This is further evidenced by other subsets, which crowd the results due to their higher lexical similarity compared to the marginalized matches. In contrast, FNs display near-zero lexical overlap, revealing the existing limitation on retrieving highly non-lexical semantic matches. Because the FNs actually exhibit the lowest Hit OOV and the most similar syntactic structure despite consistently low query-side noise, this disparity confirms a critical paradox: while the model is remarkably robust against OCR degradation and syntactic drift, its ranking remains anchored to surface-level lexical similarity, systematically marginalizing structurally aligned historical concepts, regardless of their superior textual integrity.

\begin{table}[t]
\centering
\setlength{\tabcolsep}{2pt} 
\resizebox{\columnwidth}{!}{
\begin{tabular}{l|c|c|cc|c}
\toprule
\textbf{Subset} & \textbf{\textit{N}} & \textbf{Vocab Sim.} & \textbf{Quote OOV} & \textbf{Hit OOV} & \textbf{POS Div.} \\
\midrule
Top P & 40 & 0.030 $\pm$ 0.026 & 4.11 $\pm$ 3.31 & 19.25 $\pm$ 7.88 & 0.348 $\pm$ 0.048 \\
Top N & 121 & 0.017 $\pm$ 0.020 & 1.73 $\pm$ 2.45 & 20.62 $\pm$ 7.62 & 0.362 $\pm$ 0.061 \\
Tail N & 197 & 0.012 $\pm$ 0.017 & 2.17 $\pm$ 2.57 & 21.23 $\pm$ 7.23 & 0.359 $\pm$ 0.064 \\
Tail P & 10 & 0.009 $\pm$ 0.019 & 3.44 $\pm$ 3.88 & 17.46 $\pm$ 2.62 & 0.331 $\pm$ 0.045 \\
\bottomrule
\end{tabular}
}
\caption{Linguistic feature distributions across English hit subsets. The rows correspond to defined TP: highly ranked (in top 30\%) significant hits (\textit{Paraphrase} or \textit{Meaning Match}), FP: highly ranked (in top 30\%) non-matches, TN: no-matches in the tail (top 60\%--90\%), and FN accordingly. \textbf{\textit{N}} represents the size of the subset.}
\label{tab:linguistic}
\end{table}


\section{Conclusion}
This study demonstrates that a human-in-the-loop semantic search workflow can surpass the limitations of traditional lexical matching, uncovering a significant volume of implicit reception within noisy historical corpora. By offering both a methodological proof-of-concept and an expert-curated reference dataset, our purpose is to provide feasibility evidence and a reusable evaluation framework. We plan to scale it into a systematic benchmark to evaluate dense retrieval models across broader historical and low-resource domains. 



\section*{Acknowledgments}
This research was funded by the European Union under the Horizon Europe Marie Skłodowska-Curie Actions (Grant Agreement No. 101119511) and the Academy of Finland (Grant Nos. 347706 and 347709). Computational resources were provided by CSC – IT Center for Science. Views and opinions expressed are however those of the authors only and do not necessarily reflect those of the European Union or Horizon Europe MSCA Actions. Neither the European Union nor the granting authority can be held responsible for them.

We extend our gratitude to Kira Hinderks, Iiro Tiihonen, Yann Ryan, and David Rosson for their extensive support and feedback during the data annotation phase, alongside the broader Computational History Group (COMHIS) network for their insightful discussions.

\newpage
\bibliography{latex/anthology-1,latex/anthology-2,latex/custom}




\end{document}